\documentclass[letterpaper]{article}
\usepackage{aaai25} 
\usepackage{times} 
\usepackage{helvet} 
\usepackage{courier} 
\usepackage[hyphens]{url} 
\usepackage{graphicx} 
\usepackage{graphicx} 
\usepackage{natbib} 
\usepackage{caption} 
\usepackage{lipsum}
\title{LLM for Barcodes: Generating Diverse Synthetic Data for Identity Documents}
\author{
    Hitesh Laxmichand Patel\textsuperscript{\rm 1},
    Amit Agarwal\textsuperscript{\rm 2},
    Bhargava Kumar\textsuperscript{\rm 3},
    Karan Gupta\textsuperscript{\rm 1}
    Priyaranjan Pattnayak\textsuperscript{\rm 4}
}
\affiliations{
    \textsuperscript{\rm 1}New York University\\
    \textsuperscript{\rm 2}Liverpool John Moores University\\
    \textsuperscript{\rm 3}Columbia University\\
    \textsuperscript{\rm 4}University of Washington\\ 
}

\begin{document}
\maketitle
\begin{abstract}
Accurate barcode detection  and decoding in Identity documents is crucial for applications like security, healthcare, and education, where reliable data extraction and verification are essential. However, building robust detection models is challenging due to the lack of diverse, realistic datasets an issue often tied to privacy concerns and the wide variety of document formats. Traditional tools like Faker rely on predefined templates, making them less effective for capturing the complexity of real-world identity documents. In this paper, we introduce a new approach to synthetic data generation that uses LLMs to create contextually rich and realistic data without relying on predefined field. Using the vast knowledge LLMs have about different documents and content, our method creates data that reflects the variety found in real identity documents. This data is then encoded into barcode and overlayed on templates for documents such as Driver’s licenses, Insurance cards, Student IDs. Our approach simplifies the process of dataset creation, eliminating the need for extensive domain knowledge or predefined fields. Compared to traditional methods like Faker, data generated by LLM demonstrates greater diversity and contextual relevance, leading to improved performance in barcode detection models. This scalable, privacy-first solution is a big step forward in advancing machine learning for automated document processing and identity verification.

\end{abstract}
\section{Introduction}
A barcode is a machine readable representation of data in the form of parallel lines or patterns. It is used to store information about the object to which it's attached and play a crucial role in how we manage and process information today. Barcodes make it possible to store, retrieve, and verify data quickly and accurately, which is essential in many industries. There are several types of barcodes, with the two primary categories being one-dimensional (1D) and two-dimensional (2D) barcodes. Barcodes play a crucial role in identity documents like insurance cards, Driver's license, and ID cards. They store essential personal information in a machine readable format, facilitating quick and accurate verification. These documents are issued by governments or authorized organizations to confirm someone’s identity. Barcodes in these documents help improve security, make verification faster, and allow different systems to work together more efficiently. However, building reliable models to detect barcodes in identity documents is a tough problem. One major challenge is the lack of large and diverse datasets needed to train these models. Privacy laws, like the General Data Protection Regulation (GDPR), restrict access to real-world identity documents, and because these documents often contain sensitive personal information, there are additional restrictions. Furthermore, identity documents vary a lot in design as they can have different data encoded, layouts, languages, and barcode types depending on the country or organization that issued them. This makes it hard for models trained on limited data to work well across all types of documents \cite{agarwal2024domain}.

Some existing tools attempt to generate synthetic data to bridge this gap \cite{agarwal2024synthetic,agarwal2024techniques}, but they come with limitations. For instance, tools like Faker rely on predefined fields, which don’t adapt well to the variability and complexity of real-world identity documents. This approach often results in data that lacks diversity and fails to accurately represent actual documents. For example, driver’s licenses from states like New Jersey and Texas encode different types of information, and the structure may vary not only between documents but also within a single document. As a result, models trained on such synthetic data may struggle to perform effectively. To solve these problems, we propose a new way to create synthetic datasets using LLMs. These models can generate realistic and diverse data for identity documents without relying on predefined fields. The data reflects the variety of global identity documents, including differences in culture and region, while also ensuring that privacy is protected. This data is then converted into barcodes and overlayed onto document templates, creating a dataset that looks and feels real but doesn’t include any sensitive personal information.

Our Key Contributions:

\begin{itemize}
    \item  We use LLMs to generate data that captures the diversity and complexity of identity documents worldwide. This approach avoids rigid templates and reduces the time and effort needed to create datasets.
    
    \item Models trained on our synthetic data perform better on a variety of document formats and barcode types compared to those trained on traditional synthetic datasets.
    
    \item Our method can easily adapt to new types of documents, regions, or barcode standards by making simple changes to the prompts used with LLMs. This makes it a versatile solution as identity document standards evolve.
\end{itemize}

\section{Related Work}

The development of accurate barcode detection and decoding systems has been significantly enhanced through the use of synthetic data and advanced data generation techniques. This section reviews key studies that have utilized synthetic data for barcode detection and explores the application of LLMs in synthetic data generation across various domains.

\subsection{Barcode Detection and Extraction}

Barcode detection and decoding have seen significant advancements with the integration of synthetic data and cutting-edge machine learning models. Models like YOLO (You Only Look Once) by \cite{redmon2016lookonceunifiedrealtime} and Faster R-CNN by \cite{ren2016fasterrcnnrealtimeobject} have been pivotal in advancing object detection. YOLOv5, as demonstrated by \cite{barcode-deep}, excelled in real-time detection across varied orientations and scales, while Faster R-CNN, utilized by \cite{barcode-deep}, proved effective in Ultra high resolution images. To address the scarcity of annotated datasets, synthetic data generation has become a key solution. \cite{Quenum_2021} introduced a synthetic dataset of 100,000 barcode images, significantly improving model performance in high-resolution contexts. Similarly, \cite{zharkov2020new} developed a hybrid dataset of 30,000 synthetic and 921 real barcode images, providing a robust benchmark for algorithm testing.
These advancements underscore how combining synthetic data with state-of-the-art models enhances barcode detection accuracy and robustness across diverse real-world scenarios.

\subsection{LLM assisted Synthetic Data}

Large Language Models are playing an increasingly important role in synthetic data generation, helping to address data scarcity and improve machine learning applications \cite{agarwal2024enhancing}. For instance, \cite{long2024llmsdrivensyntheticdatageneration} provides a detailed overview of LLM-driven workflows for creating synthetic data, along with the challenges they face. \cite{patel2024datadreamertoolsyntheticdata} introduces DataDreamer, a tool designed to ensure that synthetic data generation using LLMs is reproducible. Similarly, \cite{gupta2024targentargeteddatageneration} showcases TarGEN, which uses targeted prompts to enhance the quality of generated datasets. On a larger scale, \cite{ge2024scalingsyntheticdatacreation} presents Persona Hub, a framework for creating synthetic personas at scale. Lastly, \cite{veselovsky2023generatingfaithfulsyntheticdata} explores methods for generating reliable synthetic data for social science research using grounding techniques. Together, these works demonstrate the power of LLMs in producing diverse, high-quality synthetic data for a wide range of purposes
\section{Method}
This section outlines how data is generated for various types of identity documents, including Driver’s Licenses, Insurance Cards, and University IDs. Each document type is designed to include specific information tailored to its purpose. For example, Driver’s Licenses typically feature details like the holder’s name, address, license number, date of birth, and issuing state, though the exact fields can vary between states. Insurance Cards, on the other hand, include information such as the policy number, coverage dates, provider, and plan type, which often differ based on the insurance company. Similarly, University IDs contain data like the student’s name, ID number, department, and enrollment year, with variations depending on the institution. These differences highlight the need for flexible and realistic metadata generation to reflect the diversity of real-world documents.
Our approach consists of three key components: data generation using a LLM and the Faker library, barcode encoding with the pyBarcode library, and integrating these barcodes into document templates. To further enrich the dataset, we apply data augmentation techniques \cite{agarwal2024techniques}, ensuring greater diversity and realism in the generated documents.

\subsection{Data Generation using LLM}

We begin by crafting detailed prompts tailored to each document type—Driver's Licenses, Insurance Cards, and University IDs. These prompts are designed to elicit comprehensive and contextually appropriate metadata from the LLMs. For example, a prompt for generating a document data might:

\textbf{"Generate data for a Driver’s License for the state {California}, country {USA}. Based on the issuing authority, the metadata should include contextually relevant fields formatted as plain text and separated by | for barcode encoding. Ensure the data is fictional but realistic and consistent with the standards for a driver’s license in the state and country."}

\textbf{"Generate data for a Insurance Card issued by {Blue Cross Blue Shield}, country {USA}. Based on the issuing company, the metadata should include relevant fields formatted as plain text and separated by | for barcode encoding. Ensure the data is fictional but realistic and consistent with company and insurance card standards."}

\textbf{"Generate data for a University Student ID issued by {Harvard University}, country {USA}. Based on the issuing university, the metadata should include contextually relevant fields formatted as plain text and separated by | for barcode encoding. Ensure the data is fictional but realistic and consistent with university id standards in the country.}

To ensure diversity, prompts encourage the inclusion of variations in cultural, regional, and linguistic elements. We also specify that all generated data should be entirely fictional and not correspond to any real individuals to maintain privacy compliance.
Using the crafted prompts, we interact with the LLM to generate metadata entries. The LLM's understanding of global contexts allows it to produce data that reflects the diversity found in real-world documents. For instance:
To simulate realistic and diverse metadata for identity documents, we utilized the Llama 70B\cite{touvron2023llamaopenefficientfoundation}. The LLM was prompted to generate synthetic personal information that adheres to standard formats used in various countries and states, ensuring both variability and authenticity in the generated data.For instance:

\begin{itemize}
    \item Driver's Licenses: The LLM generates names, addresses, dates of birth, license numbers, issuing states or provinces, and other relevant details, varying across different regions and languages
    \item Insurance Cards: It produces policy numbers, coverage dates, provider names, plan types, and member information, reflecting different insurance providers and policies.
    \item University IDs: The LLM generates student names, ID numbers, departments, enrollment years, and university names, capturing a wide range of academic institutions and programs.  
\end{itemize}
We automate this process to generate large volumes of metadata, ensuring a broad and diverse dataset.

\subsection{Data Generation using Faker} 
While the Faker library offers tools to generate fake data, it does not natively support specific formats such as the AAMVA(American Association of Motor Vehicle Administrators) standard for driver's licenses or the varied layouts of insurance cards and college IDs across regions. To overcome this limitation, we developed custom templates tailored to the structural and formatting details of these documents.

These templates were designed for driver's licenses, insurance cards, and college IDs, incorporating regional and institutional variations. By extending the Faker library, we filled these templates with synthetic data, ensuring they adhered to the required formats. This approach allowed us to generate metadata for a wide range of identity documents.

\subsection{Barcode Creation}

Using the data from both the LLM and Faker-generated datasets, we encoded the data into barcodes appropriate for each document type using the pyBarcode library. This ensured consistency in the barcode generation process across both datasets.

Driver's Licenses (DL): We generated PDF417 barcodes, commonly used for high-density data encoding on identification cards, adhering to the AAMVA standard.
Insurance Cards: Code 128 barcodes were employed to encode alphanumeric data efficiently, matching the formats used by various insurance providers.
University IDs and Others: PDF417 is generally used for University ID.
By utilizing pyBarcode, we ensured that the barcodes conformed to industry standards, including appropriate error correction levels and encoding parameters.
\begin{figure}[h!]
    \centering
    \includegraphics[width=0.45\textwidth]{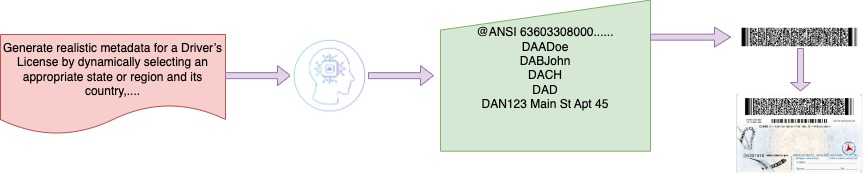}
    \caption{LLM-assisted workflow for generating synthetic driver's license data and documents.}
    \label{fig:data generation process}
\end{figure}

\subsection{Embedding Barcodes into Document Templates}

With the generated barcodes, we integrated them into customized document templates. Each template featured predefined placeholders and coordinates for precise barcode placement, while textual fields remained blank. The process for each dataset involved:

\begin{itemize}
    \item  Template Selection: Choosing the appropriate document template based on the document type, region, and institution (e.g., a driver's license template for a specific state or a university ID template for a particular institution).
    
    \item Barcode Overlay: Placing the generated barcode onto the template at the predefined coordinates corresponding to the barcode placeholder, ensuring correct alignment and sizing.
\end{itemize}

By overlaying the barcodes onto the templates using known coordinates, we created images of documents where the barcodes are accurately positioned, while the textual fields remained as placeholders or blanks. This approach focuses on the barcode data, which is essential for our training objectives, and simplifies the generation process by not rendering textual data on the templates. This meticulous integration is crucial for mimicking the spatial relationships found in real-world documents, thereby enhancing the training data's authenticity.

\subsection{Data Augmentation}

To simulate real-world variations and enhance the robustness of the trained models, we applied various data augmentation techniques to the synthesized images from both datasets. These augmentations mimic common distortions encountered in practical scenarios, such as scanning artifacts, lighting variations, and physical document wear.
The augmentation techniques we used are:

\begin{itemize}
    \item \textbf{Noise Injection:} Adding Gaussian noise to simulate sensor noise inherent in imaging devices.  
    \item \textbf{Blurring:} Applying Gaussian and motion blur to emulate out-of-focus images or movement during capture.
    \item \textbf{Color and Contrast Adjustments:} Modifying brightness, contrast, saturation, and hue to reflect different lighting conditions.
    \item \textbf{Geometric Transformations:} Introducing rotations, scaling, and perspective distortions to mimic variations in document positioning during scanning or photography.  
\end{itemize}

These augmentations were applied probabilistically to each document, promoting the development of models resilient to a wide range of real-world distortions.

\section{Experimental Setup}
To evaluate the effectiveness of our synthetic datasets for training robust models in barcode detection we conducted experiments using YOLOv5 \cite{ultralytics2021yolov5}, a state-of-the-art object detection model known for its real-time application capabilities. The goal was to compare the impact of two different data generation methods, one leveraging a LLM-based approach and the other using the Faker library with customized templates on the performance of models trained on these datasets.

We generated two separate synthetic datasets as described in the methodology section. The first dataset was created using data synthesized by the Llama 70B\cite{touvron2023llamaopenefficientfoundation} model, while the second dataset was produced using the Faker library with customized templates. Both datasets consisted of 10,000 images of driver's licenses, health cards, and university IDs, with barcodes added to the templates. The data was split into training (80\%) and validation (20\%) subsets, ensuring an even distribution of document types.

For training, we fine-tuned the YOLOv5 model to detect and classify barcode types in document images. The images were resized to 640×640 pixels to meet the model’s input requirements. Training was conducted using an initial learning rate of \(1 \times 10^{-4}\) with cosine annealing, a batch size of 16, and the Adam optimizer. Each model was trained for 10 epochs, with early stopping based on validation loss to mitigate overfitting. The same training setup was applied to both datasets for a fair comparison.
To evaluate performance, we used metrics such as mean Average Precision (mAP) at Intersection over Union (IoU) thresholds of 0.5 (mAP@0.5) and 0.75 (mAP@0.75), along with Precision, Recall, and F1-score for barcode classification. For a thorough assessment, the trained models were tested on a separate set of 750 real-world document images collected through vendors internally that were not included in the training process. This approach was critical for determining how well the models generalized from synthetic datasets to real-world applications.

The results showed that the model trained on the LLM-generated dataset consistently outperformed the one trained on the Faker-generated dataset across all evaluation metrics.

\begin{table}[h!]
\centering
\begin{tabular}{|l|c|c|}
\hline
\textbf{Metric} & \textbf{Dataset-LLM} & \textbf{Dataset-Faker} \\
\hline
mAP@0.5 & 92.5\% & 88.3\% \\
mAP@0.75 & 85.7\% & 79.4\% \\
Precision & 93.2\% & 89.1\% \\
Recall & 91.8\% & 87.6\% \\
F1-score & 92.5\% & 88.3\% \\
\hline
\end{tabular}
\caption{Performance Comparison of Dataset-LLM vs Dataset-Faker}
\label{tab:performance_comparison}
\end{table}

\section{Data Diversity}

In this section, we evaluate the diversity of our synthetic dataset generated using the LLM and the Faker library by employing two key metrics: \textbf{Unique value counts} and \textbf{Shannon Entropy} across multiple data fields.

\textbf{Unique Value Counts:}
We calculate the number of unique entries in critical data fields:
\begin{itemize}
    \item Names and Addresses: Full names of individuals and complete addresses including street, city, state, and zip code
    \item Policy Numbers,Driver License,University IDs : Insurance policy identifiers.State-specific driver license identifiers and Unique identifiers for university students
\end{itemize}
A higher count of unique values in these fields indicates greater variability and reduces the likelihood of model overfitting to specific data patterns.

\textbf{Shannon Entropy:}
Shannon entropy measures the randomness and unpredictability within categorical fields. It is calculated using the formula:
\[
H(X) = - \sum_{i=1}^{n} P(x_i) \log_2 P(x_i)
\]

where \(P(x_i)\) is the probability of occurrence of the \(i\)-th unique value. Higher entropy values reflect greater diversity and complexity in the data, which is beneficial for training robust models.

\begin{table}[h!]
\centering
\scriptsize
\begin{tabular}{|l|c|c|}
\hline
\textbf{Metric} & \textbf{Dataset-LLM} & \textbf{Dataset-Faker} \\
\hline
Unique Names & 9867 & 7234 \\
\hline
Unique Addresses & 9879 & 6569 \\
\hline
Unique Policy Numbers & 10000 & 9791 \\
\hline
Unique Driver License Numbers & 10,000 & 10,000 \\
\hline
Unique University IDs & 10,000 & 10,000 \\
\hline
Average Name Entropy (bits) & 13.2 & 10.5 \\
\hline
Average Address Entropy (bits) & 12.8 & 9.8 \\
\hline
\end{tabular}
\caption{Presents the diversity metrics for both the LLM-generated and Faker-generated datasets across the specified data fields}
\label{tab:dataset_comparison}
\end{table}

The LLM-generated dataset exhibits higher diversity compared to the Faker-generated dataset across all evaluated metrics. The LLM-generated dataset contains significantly more unique names and addresses, Policy number indicating broader variability. Both datasets have 10,000 driver license numbers, ensuring unique identifiers across documents. Higher entropy values in the LLM-generated dataset across all fields suggest a more unpredictable and diverse distribution of categorical data. This increased entropy indicates that the LLM is capable of generating more varied and complex entries compared to the Faker library.
This enhanced diversity in the LLM-generated dataset likely contributes to the superior model performance observed in Table 1. Models trained on more diverse data are better equipped to generalize to unseen, real-world data, reducing the risk of overfitting and improving robustness.

\section{Conclusion and Limitations}

This study introduces a synthetic data generation pipeline using LLMs to create diverse and realistic data for identity documents. By combining LLM-generated content with barcode encoding and document synthesis, it addresses data scarcity and variability, enhancing barcode detection models and ensuring privacy compliance across various document formats.

However, the study has limitations, as it was tested on a small set of real images, limiting its real-world evaluation. Additionally, the scalability and cost efficiency of smaller LLMs remain unexplored. Despite these challenges, this work significantly advances automated document processing and identity verification in privacy-sensitive applications.

\bibliography{aaai25}
\end{document}